%% file: main.tex
\documentclass[nohyperref]{article}

\usepackage{microtype}
\usepackage{graphicx}
\usepackage{subfigure}
\usepackage{booktabs} 

\usepackage{hyperref}

\usepackage[accepted]{icml2023}

\usepackage{amsmath}
\usepackage{amssymb}
\usepackage{mathtools}
\usepackage{amsthm}

\usepackage{multirow}

\usepackage[capitalize,noabbrev]{cleveref}

\theoremstyle{plain}

\theoremstyle{definition}

\theoremstyle{remark}

\usepackage[textsize=tiny]{todonotes}
\input{math_commands.tex}
\newcommand{\ours}{\texttt{D-MFDAL}}
\icmltitlerunning{Disentangled Multi-Fidelity Deep Bayesian Active Learning}

\begin{document}

\twocolumn[
\icmltitle{Disentangled Multi-Fidelity Deep Bayesian Active Learning}



\icmlsetsymbol{equal}{*}

\begin{icmlauthorlist}
\icmlauthor{Dongxia Wu}{yyy,zzz}
\icmlauthor{Ruijia Niu}{yyy}
\icmlauthor{Matteo Chinazzi}{kkk,aaa}
\icmlauthor{Yi-An Ma}{zzz,yyy}
\icmlauthor{Rose Yu}{yyy,zzz}
\end{icmlauthorlist}

\icmlaffiliation{yyy}{Department of Computer Science and Engineering, University of California San Diego, La Jolla, USA}
\icmlaffiliation{zzz}{Halıcıoğlu Data Science Institute, University of California San Diego, La Jolla, USA}
\icmlaffiliation{kkk}{The Roux Institute, Northeastern University, Portland, USA}
\icmlaffiliation{aaa}{Network Science Institute, Northeastern University, Boston, USA}


\icmlcorrespondingauthor{Rose Yu}{roseyu@ucsd.edu}

\icmlkeywords{Multi-fidelity surrogate modeling, Bayesian Active Learning, Neural Processes}

\vskip 0.3in
]



\printAffiliationsAndNotice{}  

\begin{abstract}
To balance quality and cost, various domain areas of science and engineering run simulations at multiple levels of sophistication. Multi-fidelity active learning aims to learn a direct mapping from input parameters to simulation outputs at the highest fidelity by actively acquiring data from multiple fidelity levels.
However, existing approaches based on Gaussian processes are hardly scalable to high-dimensional data. Deep learning-based methods often impose a hierarchical structure in hidden representations, which only supports passing information from low-fidelity to high-fidelity. These approaches can lead to the undesirable propagation of errors from low-fidelity representations to high-fidelity ones. 
We propose a novel framework called Disentangled Multi-fidelity Deep Bayesian Active Learning (\ours{}), which learns the surrogate models conditioned on the distribution of functions at multiple fidelities.
On benchmark tasks of learning deep surrogates of partial differential equations including heat equation, Poisson's equation and fluid simulations, our approach significantly outperforms state-of-the-art in prediction accuracy and sample efficiency. 
\end{abstract}

\section{Introduction}
\input{sections/intro}

\section{Background}
\input{sections/background}

\section{Methodology}
\input{sections/method}
\section{Related Work}
\input{sections/relate}

\section{Experiments}
\input{sections/exp}
\section{Conclusion}
\input{sections/conc}

\section{Acknowledgments}
This work was supported in part by U.S. Department Of Energy, Office of Science, Facebook Data Science Research Awards, U. S. Army Research Office under Grant W911NF-20-1-0334, and NSF Grants \#2134274 and \#2146343, as well as NSF-SCALE MoDL (2134209) and NSF-CCF-2112665 (TILOS). M.C. acknowledges support from grant HHS/CDC 5U01IP0001137.

\clearpage
\bibliographystyle{icml2023}
\bibliography{ref}


%


\clearpage
\appendix

\end{document}

%% file: math_commands.tex
\usepackage{amsmath,amsfonts,bm}

\def\eqref#1{equation~\ref{#1}}
\def\Eqref#1{Equation~\ref{#1}}

\def\1{\bm{1}}

\DeclareMathAlphabet{\mathsfit}{\encodingdefault}{\sfdefault}{m}{sl}
\SetMathAlphabet{\mathsfit}{bold}{\encodingdefault}{\sfdefault}{bx}{n}

%% file: sections/intro.tex
Mathematical modeling and simulations play a crucial role in various scientific and engineering fields, ranging from diffusion modeling to epidemic simulation. These models can often be simulated at different levels of sophistication. High-fidelity models provide highly accurate results but require more computational resources, while low-fidelity models offer less accuracy but are less computationally expensive. Multi-fidelity modeling, as outlined in \cite{peherstorfer2018survey}, aims to strike a balance between computation cost and prediction accuracy by using data from multiple levels of fidelity to learn an accurate high-fidelity surrogate. The learned surrogate can replicate the behavior of the original model to eliminate the complex numerical integration.

While Gaussian processes (GPs) remain to be predominant tools in multi-fidelity modeling \citep{perdikaris2016multifidelity,wang2021multi}, deep learning arises as a more scalable alternative for high-dimensional data \cite{cutajar2019deep,wang2020mfpc,hebbal2021multi, wu2022multi}. These methods use a deep neural network to learn a direct mapping from input parameters to simulation outputs using multi-fidelity data. However, they also require simulating massive training data beforehand, which is expensive to obtain, especially for high-fidelity simulation.

Multi-fidelity deep active learning (MFDAL) \cite{li2022deep, li2022batch} proposes a framework to acquire data at different fidelity levels with deep learning and to reduce the cost of data simulation. Such models pass information from low-fidelity to high-fidelity hidden representations through a neural network (NN). This design requires accurate hidden representations at each fidelity to propagate useful information from low-fidelity to high-fidelity levels. However,  in multi-fidelity active learning, these hidden representations can be easily erroneous when the number of training data is highly unbalanced at each fidelity and the data distribution is dramatically shifted during the beginning stage of active learning. Moreover, the trained surrogate model will also have the overfitting issue at the beginning stage with limited training data at each fidelity level. These overfitted hidden representations are less accurate and their error will propagate from low-fidelity to high-fidelity. 


To alleviate the overfitting problem, \cite{wu2022multi} propose a unified neural latent variable model for multi-fidelity surrogate modeling called Multi-fidelity Hierarchical Neural Processes (MFHNP). They introduce latent variables to learn the distributions over functions at each fidelity level. However, this model still requires a hierarchical structure to pass information from low-fidelity to high-fidelity levels via hidden representations of a NN. Therefore, the error propagation issue remains. 



In this work, we design a novel framework called Disentangled Multi-fidelity Deep Bayesian Active Learning (\ours{}) to learn the multi-fidelity representations in the functional space. \ours{} is able to solve both error propagation and overfitting issues mentioned above.  Specifically, \ours{} belongs to the Neural Process (NP) family \cite{garnelo2018neural,garnelo2018conditional} to learn the latent variables from the individual latent representations of the input-output pairs in the context set. The latent variables are used to represent the distributions over functions at each fidelity level. \ours{} disentangles these individual latent representations into two parts for global-local separation. The global representations are treated as the samples generated from latent representations among all fidelity levels, while the local ones are samples generated from latent representations at individual fidelity level. In this way, \ours{} avoids the hierarchical model architecture.

We design a unified evidence lower bound (ELBO) for the joint distribution among all fidelity levels as the training loss and introduce the multi-fidelity regularization term to enforce similar global representations across the fidelity levels for the same sample. Furthermore, we extend the acquisition function, latent information gain \cite{wu2021accelerating}, designed for Bayesian active learning on NP-based models to multi-fidelity setting and design an efficient algorithm for budget-constrained batch active learning.


In summary, our contributions include:
\begin{itemize}
    \item A scalable Disentangled Multi-fidelity Deep Bayesian Active Learning framework (\ours{}). The disentangled representation makes it flexible and efficient to share global information across all fidelity levels.
    \item  A novel acquisition function called Multi-fidelity Latent Information Gain (MF-LIG) and an efficient algorithm for budget-constrained greedy-based batch active learning implementation.
    \item Superior performance in multiple benchmark studies of learning deep surrogates of partial differential equations and complex fluid prediction task in both passive learning and active learning settings.
\end{itemize}

\begin{figure*}[h]
    \centering
    \includegraphics[width=\linewidth,trim={0, 0, 0, 0}]{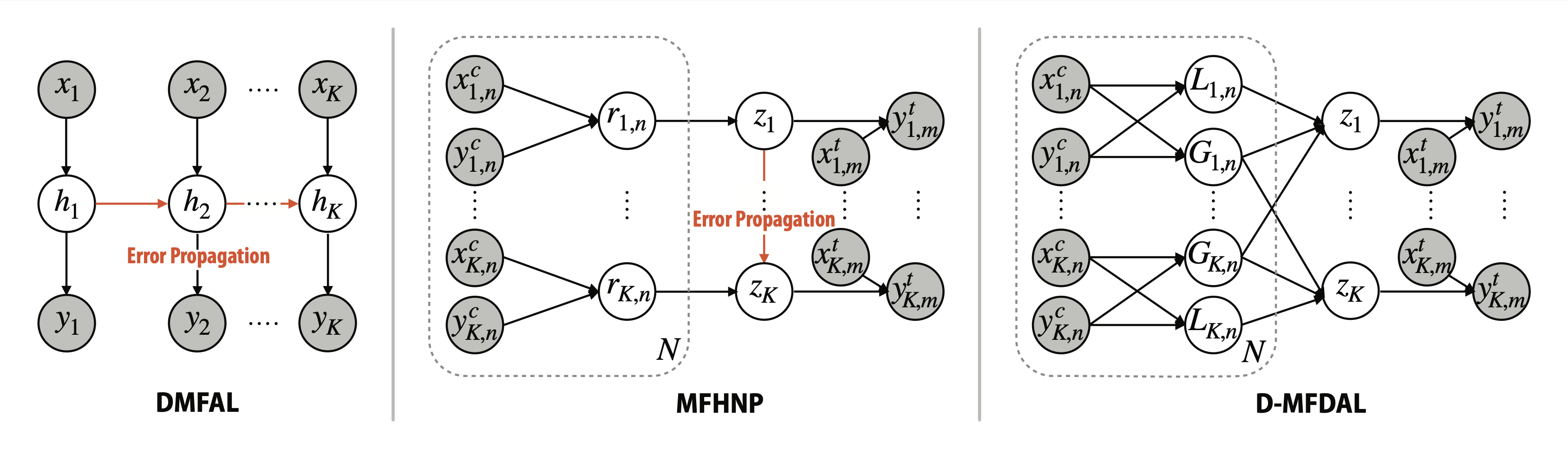}
    \caption{Graphical model: Left and Middle: two multi-fidelity surrogate modeling baselines. Both have hierarchical structures. They use the hidden variable $h_k$ or the latent variable $z_k$ to pass information from low-fidelity to high-fidelity levels and therefore suffer from the error propagation issue. Right: \ours{} disentangles the latent representations $r_{k,n}$ shown in MFHNP into local representations $L_{k,n}$ and global representations $G_{k,n}$, and directly uses them to infer the latent variable $z_k$. $z_k$ are conditionally independent of each other given the local and global representations. Shaded circles denote observed variables and hollow circles represent latent variables. The directed edges represent conditional dependence.
    }
    \label{fig:fig1}
\end{figure*}

%% file: sections/background.tex
\paragraph{Muti-Fidelity Modeling.} Formally, given input domain $  \mathcal{X} \subseteq  \mathbb{R}^{d_x}$ and output domain $ \mathcal{Y} \subseteq  \mathbb{R}^{d_y}$,   a model is a (stochastic) function $f:\mathcal{X} \rightarrow \mathcal{Y}$.  The evaluations of $f$ incur computational costs $c> 0$. The computational costs $c$ are higher at higher fidelity level ($c_1 <... < c_K $). 
In multi-fidelity modeling, we have a set of functions $\{f_{1},\cdots, f_{K}\}$ that approximate $f$ with increasing accuracy and computational cost. Our target is to learn a deep surrogate model $\hat{f}_{K}$ based on data from $K$ fidelity levels and $N$ different parameter settings (scenarios) $\{x_{k,n}, y_{k,n}\}_{k=1,n=1}^{K,N}$.

\paragraph{Neural Processes.} Neural processes (NPs) \citep{garnelo2018neural} are a family of conditional latent variable models for implicit stochastic processes ($\mathcal{SP}s$) \citep{wang2020doubly}. NPs combine GPs and neural networks (NNs). Like GPs, NPs can represent distributions over functions and can estimate the uncertainty of the predictions. But they are more scalable in high dimensions and allow continual and active learning out-of-the-box \citep{jha2022neural}. 

%
%

According to Kolmogorov Extension Theorem \citep{oksendal2003stochastic}, NPs meet exchangeability and consistency conditions to define $\mathcal{SP}s$. Formally, NP includes latent variables $z \in \mathbb{R}^{d_z}$ and model parameters $\theta$ and is trained by the context set $\mathcal{D}^c \equiv \{x^c_{n},y^c_{n}\}_{n=1}^{N}$ and target sets $\mathcal{D}^t \equiv \{x^t_{m},{y^t_{m}}\}_{m=1}^{M}$. Here $\mathcal{D}^c$ and $\mathcal{D}^t$ are randomly split from the training set $\mathcal{D}$. Learning the posterior of $z$ and $\theta$ is equivalent to maximizing the following posterior likelihood:
%
\begin{equation}
p(y^t_{1:M}|x^t_{1:M},\mathcal{D}^c,\theta) = \int p(z|\mathcal{D}^c,\theta)\prod^{M}_{m=1}p(y^t_{m}|z, x^t_{m},\theta)dz 
\label{eqn:np}
\end{equation}
%
%
Since marginalizing over the latent variables $z$ is intractable, the NP family \citep{garnelo2018neural, kim2019attentive} uses approximate inference and derives the corresponding evidence lower bound (ELBO):

\vskip -0.12in
\begin{align}
    & \log p(y^t_{1:M}|x^t_{1:M},\mathcal{D}^c,\theta) \geq \nonumber\\ 
    & \mathbb{E}_{q_\phi(z|\mathcal{D}^c \cup \mathcal{D}^t)} \big[  \sum_{m=1}^M\log p(y^t_m|z, x^t_m,\theta)+\log\frac{q_\phi(z|\mathcal{D}^c)}{q_\phi(z|\mathcal{D}^c \cup \mathcal{D}^t)}\big] 
\end{align}

Note that this variational approach approximates the intractable true posterior $p(z|\mathcal{D}^c, \theta)$ with the approximate posterior $q_\phi(z|\mathcal{D}^c)$. This approach is also an amortized inference method as the global parameters $\phi$ are shared by all context data points. It is efficient during the test time (no per-data-point optimization) \citep{volpp2020bayesian}.

%% file: sections/method.tex
\begin{figure}[t!]
    \centering
    \includegraphics[width=0.82\linewidth,trim={0, 0, 0, 0}]{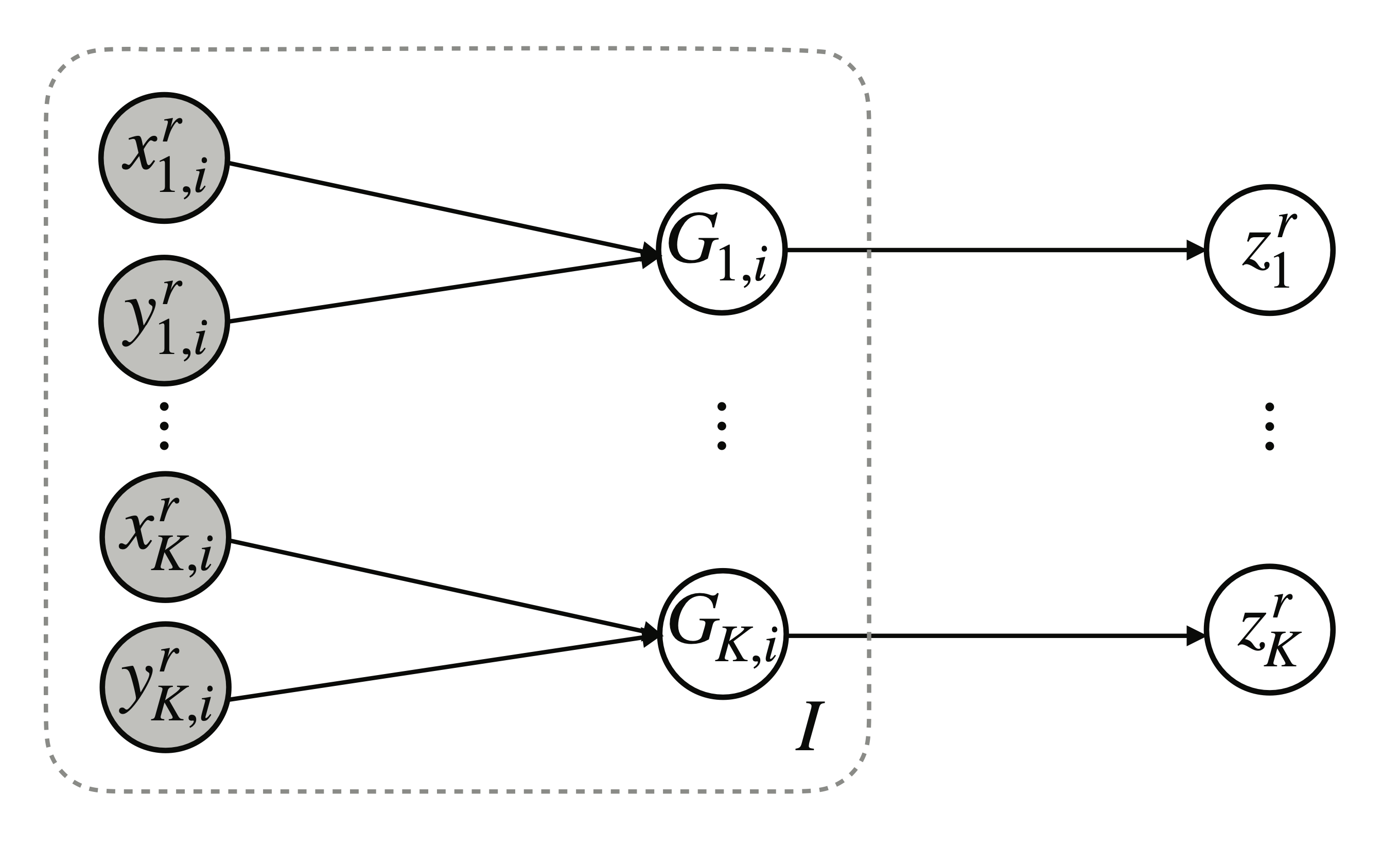}
    \caption{Graphical model: Inference graph for the reference context pairs $\{x_{k,i}^{r}, y_{k,i}^{r}\}$. Shaded circles denote observed variables and hollow circles represent latent variables. The directed edges represent conditional dependence.
    }
    \label{fig:fig2}
\end{figure}

Our proposed \ours{} is presented in two sections. First, we describe the disentangled neural processes architecture, specifically designed for multi-fidelity surrogate modeling and the associated training procedure. Secondly, we introduce a new acquisition function (MF-LIG) for multi-fidelity active learning, which extends Latent Information Gain \cite{wu2021accelerating}. Additionally, we present a greedy-based algorithm for batch active learning under budget constraints.

\subsection{Disentangled Multi-fidelity Neural Processes}
We design a NP based model, Disentangled Multi-fidelity Neural Processes (DMFNP), to efficiently integrate information from multiple fidelity levels without the hierarchical structure. 

 \paragraph{Local and Global Latent Representations.} The key idea of the \ours{} model  is to disentangle latent representations $r_{k,n}$ into local representations $L_{k,n}$ and global representations $G_{k,n}$, see Figure \ref{fig:fig1} right. Intuitively, $G_{k,n}$ embeds the information from the context pair $\{x^c_{k,n}, y^c_{k,n}\}$ that can be shared to all fidelity levels, where $k$ is the fidelity level of the context pair and $n$ is the scenario index. On the other hand, $L_{k,n}$ embeds the information from the context pair $\{x^c_{k,n}, y^c_{k,n}\}$ that is only for the fidelity level $k$. 


\paragraph{Multi-fidelity Bayesian Context Aggregation.} We extend Bayesian aggregation (BA) \cite{volpp2020bayesian} to infer latent variables $z_k$. We learn the local and global representation $L_{k,n}$, $G_{k,n}$ together with the corresponding variance $\sigma^2_{L_{k,n}}$, $\sigma^2_{G_{k,n}}$. The local representation $L_{k,n}$ can be considered as a sample of $p(z_k)$. On the other hand,  we treat the global representation $G_{k,n}$  as $K$ copies of samples of $p(z_k)$ across all fidelity levels. Then we aggregate local and global representations of context data pairs to infer $z$ following the graph in Figure \ref{fig:fig1}. We implement it using the factorized Gaussian observation model with the following form:
\begin{eqnarray}
p(L_{k,n}|z_{k}) =& \mathcal{N}(L_{k,n}|z_{k}, \text{diag}(\sigma^2_{L_{k,n}})),\nonumber\\ 
 L_{k,n} =& \text{enc}_{\phi}(x^C_{k,n},y^C_{k,n}).\nonumber\\
 p(G_{k,n}|z_{k}) =& \mathcal{N}(G_{k,n}|z_{k}, \text{diag}(\sigma^2_{G_{k,n}})),\nonumber\\ 
 p(G_{k,n}|z_{m}) =& \mathcal{N}(G_{k,n}|z_{m}, \text{diag}(\sigma^2_{G_{k,n}})), \text{for all } m \neq k \nonumber\\ 
 G_{k,n} =& \text{enc}_{\phi}(x^C_{k,n},y^C_{k,n}).
\end{eqnarray}
We use factorized Gaussian priors 
\[p_0 (z_k) := \mathcal{N}(z_k|\mu_{z_{k,0}}, \text{diag}(\sigma^2_{z_{k,0}}))\] 
to derive a multi-fidelity Gaussian aggregation model and update the parameters of the posterior distribution $q_\phi (z_k|\mathcal{D}^c)$ in a closed form:
\begin{align}
 &\sigma^2_{z_{k}} = \big[(\sigma^2_{z_{k,0}})^\ominus + \sum^N_{n=1}(\sigma^2_{L_{k,n}})^\ominus) + \sum^K_{j=1}[\sum^N_{n=1}(\sigma^2_{G_{j,n}})^\ominus)] \big]^\ominus, \nonumber \\
&\mu_{z_{k}} = \mu_{z_{k,0}} + \sigma^2_{z_{k}}\odot \big[\sum^N_{n=1}(L_{k,n} - \mu_{z_{k,0}}) \oslash (\sigma^2_{L_{k,n}}) \nonumber \\
& \:\:\:\:\:\:\:\:+ \sum^K_{j=1}[\sum^N_{n=1}(G_{j,n} - \mu_{z_{k,0}})\oslash(\sigma^2_{G_{j,n}})]\big]. 
\end{align}

where $\ominus$, $\odot$ and $\oslash$ denote element-wise inversion, product, and division, respectively.

\paragraph{Unified ELBO.} We design a unified ELBO based on the \ours{} model. For multi-fidelity surrogate modeling, we  infer the latent variables $z_k$ at each fidelity level. Therefore, we use $K$ encoders $q_{\phi_k}(z_k|\mathcal{D}^c)$ and $K$ decoders $p_{\theta_k}(y^t_{k}|z_k,x^t_k)$ for $k \in \{1,...,K\}$. When $K=2$, we can derive the corresponding ELBO containing $4$ terms as:




\begin{align}
    & \log p(y^t_{1},y^t_{2}|x^t_{1},x^t_{2},\mathcal{D}^c,\theta) \nonumber\\
\geq&  \mathbb{E}_{q_\phi(z_1,z_2|\mathcal{D}^c \cup \mathcal{D}^t)} \big[\log p(y^t_1,y^t_2|z_1,z_2, x^t_1, x^t_2,\theta)  + \nonumber\\
    & \log \frac{q_\phi(z_1,z_2| \mathcal{D}^c)}
    {q_\phi(z_1,z_2|\mathcal{D}^c \cup \mathcal{D}^t)} \big]\nonumber\\
    =& \mathbb{E}_{q_{\phi_2}(z_2|\mathcal{D}^c \cup \mathcal{D}^t)q_{\phi_1}(z_1|\mathcal{D}^c \cup \mathcal{D}^t)}\big[ \log p(y^t_2|z_2, x_2^t,\theta_2) + \nonumber\\
 & \log p(y_1^t|z_1, x_1^t,\theta_1) + \log \frac{q_{\phi_2}(z_2|\mathcal{D}^c)}{q_{\phi_2}(z_2|\mathcal{D}^c \cup \mathcal{D}^t)} +\nonumber\\
  & \frac{q_{\phi_1}(z_1|\mathcal{D}^c)}{q_{\phi_1}(z_1|\mathcal{D}^c \cup \mathcal{D}^t)}\big] 
    \label{eqn:elbo}
\end{align}

Such a unified ELBO objective can be generalized to accommodate any desired number of fidelity levels.

\paragraph{Multi-Fidelity Regularization.} Since $G_{k,n}$ is the global representation, any $(G_{k_1,i}, G_{k_2,i})$ pair should be similar across fidelity levels for the same scenario $i$. However, since the output dimensions are different at each fidelity level, \ours{} cannot share the encoder at different fidelity levels. Therefore, we introduce reference context data $\mathcal{D}_k^r=\{x_{k,i}^{r}, y_{k,i}^{r}\}_{i=1}^I$, which is shared across all fidelity levels (see Figure \ref{fig:fig2} for the inference graph). $I$ is the total number of reference scenarios. We design the multi-fidelity regularization term to minimize the Jensen–Shannon divergence between the inferred posterior $z_k^{r}$ distribution from $(x_{k,i}^{r}, y_{k,i}^{r})$ pairs (where $k < K$) and the posterior $z_K^{r}$ distribution from $(x_{K,i}^{r}, y_{K,i}^{r})$ pairs. 
Note that \ours{} does not require additional data  as we use the initial training data as reference data for fair comparison. 

We use factorized Gaussian priors for reference latent representations $z_k^{r}$:
\[p_0 (z_k^{r}) := \mathcal{N}(z_k^{r}|\mu_{z_{k,0}^{r}}, \text{diag}(\sigma^2_{z_{k,0}^{r}}))\]
The posterior distribution $q_\phi (z_k^{r}|\mathcal{D}_k^r)$ can be written as:

\begin{align}
    & \sigma^2_{z_{k}^{r}} = \big[(\sigma^2_{z_{k,0}^{r}})^\ominus + \sum^N_{n=1}(\sigma^2_{G_{k,n}})^\ominus) \big]^\ominus, \nonumber \\
    & \mu_{z_{k}^{r}} = \mu_{z_{k,0}^{r}} + \sigma^2_{z_{k}^{r}}\odot \big[\sum^N_{n=1}(G_{k,n} - \mu_{z_{k,0}^{r}}) \oslash (\sigma^2_{G_{k,n}}) \big].
\end{align}

We further derive the multi-fidelity regularization using the sum of Jensen-Shannon divergence between the highest fidelity level $K$ and all other lower fidelity levels $k$ as:
\begin{align}
    & \sum_{k=1}^{K} \text{JSD}(q_\phi (z_k^{r}|\mathcal{D}_k^r), q_\phi (z_K^{r}|\mathcal{D}_K^r)) \nonumber\\
    =&  \frac{1}{2} \sum_{k=1}^{K} \mathbb{E}_{q_\phi (z_k^{r}|\mathcal{D}_k^r)}\big[\log \frac{q_\phi (z_K^{r}|\mathcal{D}_K^r)}{q_\phi (z_k^{r}|\mathcal{D}_k^r)}\big]\nonumber\\
    \;\;\;+& \frac{1}{2} \sum_{k=1}^{K} \mathbb{E}_{q_\phi (z_K^{r}|\mathcal{D}_K^r)}\big[\log \frac{q_\phi (z_k^{r}|\mathcal{D}_k^r)}{q_\phi (z_K^{r}|\mathcal{D}_K^r)}\big] 
    \label{eqn:regul}
\end{align}



\paragraph{Training Procedure.} \ours{} is designed for scalable training, which means the model inference time should scale at most linearly with respect to the number of fidelity levels. It can be realized by using the disentangled latent representations to share the information across the fidelity levels. In this way, the latent variables $z_k$ are conditionally independent to each other given the global representations $G$ and the local representations $L$. Therefore, we no longer require nested Monte Carlo (MC) sampling of $z_k$ from low-fidelity to high-fidelity levels as in previous models with hierarchical structures.

For the training loss including ELBO in \Eqref{eqn:elbo} and multi-fidelity regularization in \Eqref{eqn:regul}, we use MC sampling to optimize the following objective function:

%
\begin{eqnarray}
    \mathcal{L}_{MC} & = \sum^{K}_{k=1}\bigg[\frac{1}{S}\sum^{S}_{s=1}\log p(y_k^t|x_k^t,z_k^{(s)}) \nonumber\\
    & - \text{KL}[q(z_k|,\mathcal{D}^c, \mathcal{D}^t))\|p(z_k|\mathcal{D}^c)] \nonumber\\
    & + \text{JSD}(q (z_k^{r}|\mathcal{D}_k^r), q (z_K^{r}|\mathcal{D}_K^r))]\bigg]
    \label{eqn:loss}
\end{eqnarray}
where the latent variables $z_k^{(s)}$ is sampled by $q_{\phi_1}(z_k|\mathcal{D}^c)$. 
The sampling time scales linearly w.r.t. the number of fidelity levels.

\begin{figure*}[t!]
    \centering
    \includegraphics[width=\linewidth,trim={0, 20, 0, 20}]{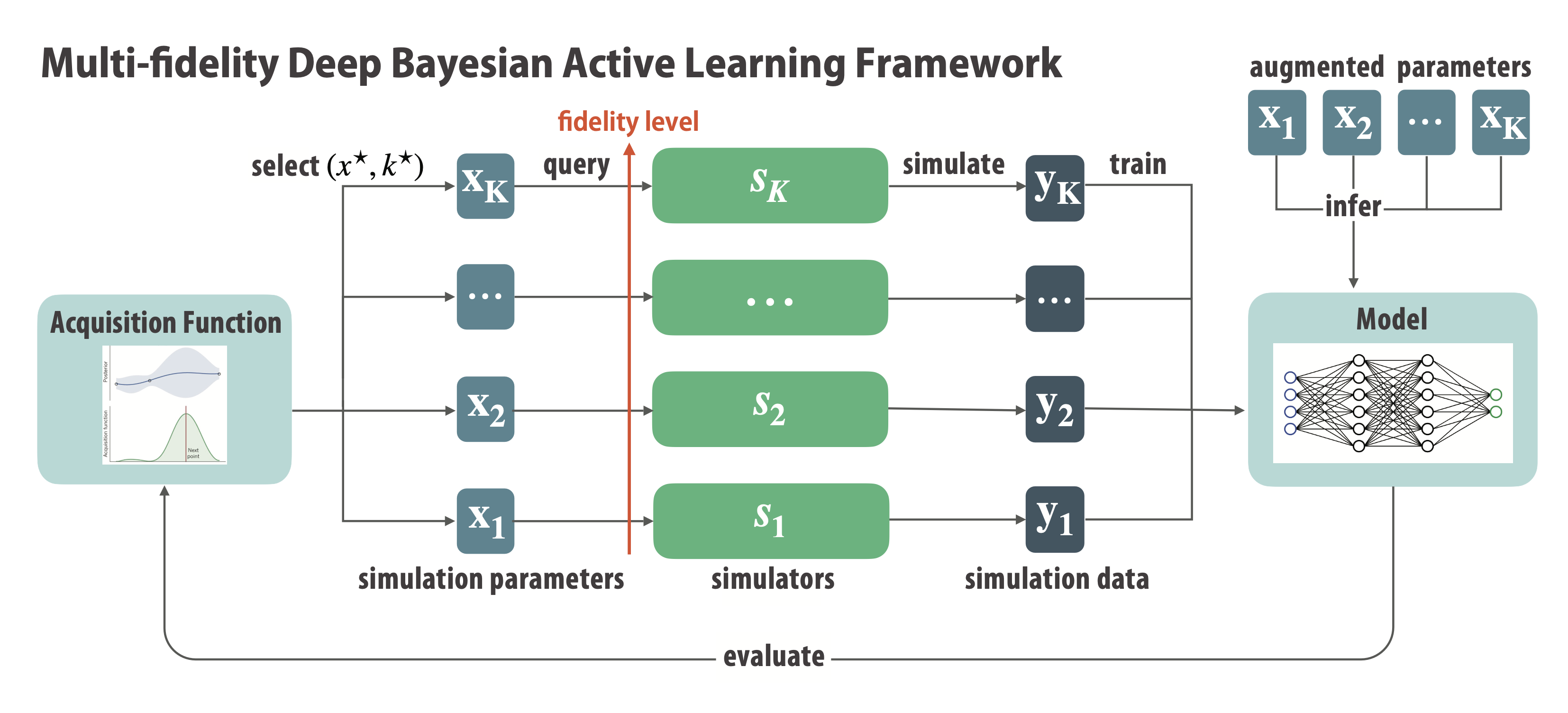}
    \caption{Illustration of the multi-fidelity deep Bayesian active learning framework (\ours{}). Given simulation parameters and data, \ours{} trains a deep surrogate model to infer the latent variables at each fidelity level. The inferred latent variables allow prediction and uncertainty quantification. The uncertainty is used to calculate the acquisition function (e.g. MF-LIG) to select the next set of parameters to query and simulate more data to add to the training set.
    }
    \label{fig:overview}
\end{figure*}

\subsection{Multi-Fidelity Active Learning}
In this section, we propose the novel acquisition function MF-LIG based on the model architecture of \ours{} for multi-fidelity active learning. Furthermore, we design a greedy batch multi-fidelity active learning algorithm with budget constraints for data efficiency.

\paragraph{Weighted Information Gain (IG). } Define the search space as $ \textit{S} = \{(x_{k,n}, y_{k,n})\}_{k=1,n=1}^{K,N}$ with $K$ fidelity levels and $N$ input parameters for each fidelity.  We flatten the  search space and define the acquisition function as:

\begin{align}
\text{IG}(x_{k,n}, y_{k,n}) = \frac{1}{c_k} [H(w ) -  H(w| x_{k,n},y_{k,n})]
\end{align}
where $c_k$ is the computational cost for level $k$. This is a naive implementation of IG for Bayesian active learning. 
In this paper, we study the continuous input parameter and discrete fidelity level setting:
\begin{eqnarray}
\text{IG}(y_{k}(x_{k})) = \frac{1}{c_k} [H(w ) - H(w| y_{k}(x_{k}))].
\end{eqnarray}

In practice, we do not know $y_{k}(x_{k})$ before querying the simulator. The best we can do is to use the weighted information gain (EIG) to replace the weighted IG:
%
\begin{align}
\text{EIG}(x_{k}) = & \frac{1}{c_k} \mathbb{E}_{p(y_{k}(x_{k}))}[H(w ) - H(w|  y_{k}(x_{k})].
\end{align}




\paragraph{Latent Information Gain for Multi-Fidelity Active Learning.} For multi-fidelity active learning, our goal is to improve the model performance at the highest fidelity level. Therefore, weighted IG/EIG is suboptimal as it treats all the model parameters $w$ at each fidelity level equally important. To find the optimal solution, we design a new acquisition function called Multi-Fidelity Latent Information Gain (MF-LIG).


We start by searching for an $x_{k}$ to optimize the EIG with respect to the model parameters used at the highest fidelity level. We can write the corresponding acquisition function:
\begin{align}
\text{MF-EIG}(x_{k}) = & \frac{1}{c_k} \mathbb{E}_{p(y_{k}(x_{k}))}[H(w_{K}) - H(w_{K}| y_{k}(x_{k}))].
\end{align}
where $w_K$ are the model parameters at the fidelity level $K$. 


The next step is to use the inferred latent variable $z_{k}$ of \ours{} to replace $w_k$ as they are learned from the context set $\{x_{k,n}^c, y_{k,n}^c\}^N_{n=1}$ to represent $f_k(.)$ of the ground truth simulators and are capable of performing conditional modeling $p(y_{k,m}^t(x_{k,m}^t)|z_k)$ at each fidelity level $k$.
We then propose a new acquisition function MF-LIG measuring the weighted expected information gain between the prediction and the latent variables at the highest fidelity level:
%
%
%
%
%
%
\begin{align}
a_s(x_k) &= \text{MF-LIG}(x_{k}) \nonumber\\
& =    \frac{1}{c_k} \mathbb{E}_{p(y_{k}(x_{k}))} \text{KL} [p(z_{K}|y_{k}(x_{k})) \| p(z_{K}))].
\label{eqn:single_acq}
\end{align}

\paragraph{Batch Multi-Fidelity Active Learning Algorithm.} We follow the greedy active learning algorithm by \cite{li2022batch} using our proposed MF-LIG for budget-constrained batch active learning. Since MF-LIG is also a mutual information based acquisition function, the guaranteed near $(1-1/e)$ approximation for the greedy algorithm also applies in our case.  Our approach is summarized in Algorithm \ref{algo:batch} and the overall framework is visualized in Figure \ref{fig:overview}.


 

\begin{algorithm}[t!]
\begin{algorithmic}
\STATE {\bfseries Input:} costs $\{c_1,...,c_K\}$, budget $B$, training set $\mathcal{D}$.
\STATE Initialize the current selected data index $j\leftarrow 0$, selected data set $\mathcal{D}^q_j \leftarrow \emptyset$, current cost $C_j \leftarrow 0$.
\WHILE{$C_j \le B$}
\STATE $(x^*, k^*) = \text{argmax}_{(x, k)} \text{MF-LIG}(x_k)$

\STATE $j \leftarrow j +1$
\STATE $\mathcal{D}^q_j \leftarrow \mathcal{D}^q_{j-1} \cup \{((x^{*}, k^{*}, \hat{y}(x^{*}, k^{*}))\} $
\STATE $\mathcal{D} \leftarrow \mathcal{D} \cup \mathcal{D}^q_j$
\STATE $C_j  \leftarrow C_{j-1}  + c_{k^{*}}$
\ENDWHILE
\STATE Return $\mathcal{D}^q_j$


\end{algorithmic}
 \caption{Batch MF-LIG}
 \label{algo:batch}

\end{algorithm}

%% file: sections/relate.tex
\paragraph{Multi-fidelity Modeling.}
Multi-fidelity surrogate modeling is widely used in science and engineering fields, from aerospace systems \cite{brevault2020overview} to climate science \cite{Hosking2020,valero2021multifidelity} \cite{valero2021multifidelity}. 
The pioneering work of \cite{kennedy2000predicting} uses GPs to relate models at multiple fidelity with an autoregressive model. \cite{le2014recursive} proposed recursive GP with a nested structure in the input domain for fast inference.  \cite{perdikaris2015multi, perdikaris2016multifidelity} deals with high-dimensional GP settings by taking the Fourier transformation of the kernel function. \citep{perdikaris2017nonlinear} proposed multi-fidelity Gaussian processes (NARGP) but assumes a nested structure in the input domain to enable a sequential training process at each fidelity level.  
\citet{wang2021multi} proposed a Multi-Fidelity High-Order GP model to speed up the physical simulation. They extended the classical Linear Model of Coregionalization (LMC) to nonlinear case and placed a matrix GP prior on the weight functions. 
Deep Gaussian processes (DGPs)  \citep{cutajar2019deep} design a single objective to optimize kernel parameters at each fidelity level jointly. However, DGPs are not scalable for applications with high-dimensional data. 

Deep learning has been applied to multi-fidelity modeling. For example, \cite{guo2022multi} uses deep neural networks to combine parameter-dependent output quantities. \cite{meng2020composite} propose a composite neural network for multi-fidelity data from inverse PDE problems. \cite{meng2021multi} propose Bayesian neural nets for multi-fidelity modeling. \cite{de2020transfer} use transfer learning to fine-tune the high-fidelity surrogate model with the deep neural network trained with low-fidelity data.    \cite{cutajar2019deep,hebbal2021multi} propose deep GPs to capture nonlinear correlations between fidelities, but their method cannot handle the case where different fidelities have data with different dimensions.  Tangentially, multi-fidelity methods have also recently been investigated in  Bayesian optimization, active learning and bandit problems \citep{li2020multi,li2022batch,perry2019allocation,kandasamy2017multi}.

Neural Processes (NPs) \cite{garnelo2018conditional, kim2018attentive, louizos2019functional,singh2019sequential} provide scalable and expressive alternatives than GPs for modeling stochastic processes. It lies between GPs and NN. However, none of the existing NP models can efficiently incorporate multi-fidelity data.  Previous work by \cite{raissi2016deep} combines multi-fidelity GP with deep learning by placing a GP prior on the features learned by deep neural networks. Their model, however, remains closer to GPs. Quite recently, \cite{wang2020mfpc} proposed multi-fidelity neural process with physics constraints (MFPC-Net). They use NP to learn the correlation between multi-fidelity data by mapping both the input and output of the low-fidelity model to high-fidelity model output. But their model requires paired data and cannot utilize the remaining unpaired data at the low-fidelity level.

\paragraph{Bayesian Active Learning.} Bayesian active learning is well studied in statistics and machine learning \citep{chaloner1995bayesian, cohn1996active}. GPs are popular for posterior estimation, e.g. \citep{houlsby2011bayesian, zimmer2018safe}, but often struggle in high dimension. Deep neural networks provide scalable solutions for active learning. Deep active learning has been applied to discrete problems such as image classification \citep{gal2017deep} and sequence labeling \citep{siddhant2018deep}. The data are queried based on different types of acquisition functions, such as predictive entropy and Bayesian Active Learning by Disagreement (BALD) \cite{houlsby2011bayesian}. \citet{kirsch2019batchbald} further developed BatchBALD, a greedy approach that incrementally selects a set of unlabeled images based on BALD score to issue batch queries for active learning. This batch acquisition function based on BALD is submodular, and therefore its corresponding greedy approach achieves a $1 - \frac{1}{e}$ approximation. Similarly, \citep{li2020deep} propose the optimization-based method DMFAL which is optmization-based and supports mutli-fidelity surrogate modeling, and BMFAL \cite{li2022batch} uses greedy approach to further extend DMFAL to support batch active-learning.






%% file: sections/exp.tex
\begin{table*}[h]
\centering
\begin{tabular}{l|l|l|l|l|l}
\toprule
Task & Setting & DMFAL & NARGP & MFHNP & \ours \\ \hline
{\multirow{3}{*}{Heat 2}} & Nested  & 0.177±2.94e-6    & 0.313 ±3.47e-6    & 0.115±8.34e-5     & \textbf{0.1  ±4.92e-5}\\ 
  & Non-nested  & 0.170±1.21e-6     & 0.311±1.71e-7    & 0.078±1.02e-4     & \textbf{0.04 ±6.4e-9}    \\ 
  & Full      & 0.138 ± 4.0e-8      & 0.31±2.12e-6    & 0.026±4.01e-5     & \textbf{0.015±1.42e-5}     \\ \hline

\hline
{\multirow{3}{*}{Heat 3}} & Nested    & 0.173 ± 1.6e-7     & 0.311±2.56e-6    & 0.145±5.11e-5    & \textbf{0.13 ±2.32e-5}     \\ 
& Non-nested  & 0.162±2.35e-6     & 0.31 ±1.05e-6    & 0.152±8.86e-5     & \textbf{0.112±2.06e-5}     \\ 
& Full      & 0.137±1.23e-7     & 0.309±3.46e-6    & 0.111±4.82e-6      & \textbf{0.108±4.85e-8}     \\ 
\hline

\hline
{\multirow{3}{*}{Poisson 2}} & Nested    & 0.179 ± 3.9e-7     & 0.595±8.71e-8    & 0.107±7.07e-5     & \textbf{0.097±5.63e-5}      \\ 
& Non-nested  & 0.157±4.56e-5     & 0.596±1.74e-5    & 0.102±4.25e-4     & \textbf{0.084±5.74e-4}     \\ 
& Full      & 0.107 ± 6.58e-5     & 0.585±9.84e-5    & 0.093±2.55e-4     & \textbf{0.07 ±2.99e-4}     \\ \hline

\hline
{\multirow{3}{*}{Possion 3}} & Nested    & 0.177±3.99e-5     & 0.594±6.3e-6    & 0.281±2.85e-5     &  \textbf{0.126±1.03e-5}      \\ 
& Non-nested  &  \textbf{0.129±6.51e-5}     & 0.592±3.77e-5    & 0.317±8.67e-5     & 0.131±3.22e-5     \\ 
& Full      & 0.121±1.47e-5     & 0.58 ±1.02e-4    & 0.335±2.37e-5     &  \textbf{0.101±1.81e-4} \\ \hline


\hline
{\multirow{3}{*}{Fluid}} & Nested    & 0.294±8.02e-8     & 0.358±1.26e-3    & 0.26 ±1.11e-6    & \textbf{0.21 ±5.13e-6}      \\ 
& Non-nested  & 0.331±6.86e-7     & 0.371±2.41e-3    & 0.263±1.67e-5     & \textbf{0.237±3.14e-6}     \\ 
& Full      & 0.275±4.59e-7     & 0.353±9.28e-4    & 0.234±4.82e-6     & \textbf{0.207±1.31e-5}     \\ \bottomrule
\end{tabular}
\caption{Passive learning performance (nRMSE) comparison of 4 different methods applied to the Heat and Poisson simulators with two and three fidelities and fluid simulation with Navier-Stokes equation. Each set of data is restructured into three settings to mimic different stages during active learning.}
\label{tb:offline}
\end{table*}

\begin{figure*}[h!]
    \centering
    \includegraphics[width=1.\linewidth,trim={0 20 0 20}]{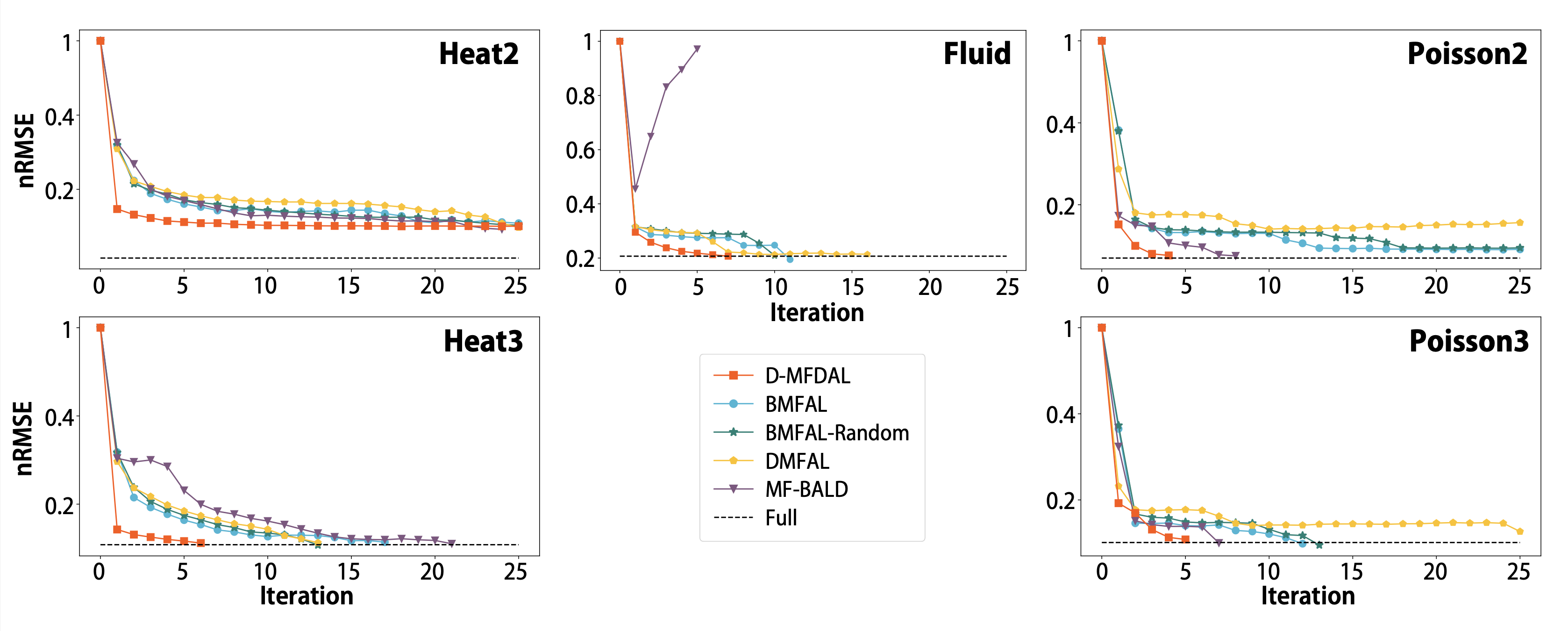}
    \caption{Active learning performance comparison for Heat and Poisson simulation with two and three fidelity levels, fluid simulation with two fidelity levels using Navier-Stokes equation. Performance is measured at the highest fidelity level.
    }
    \label{fig:fig3}
\end{figure*}

\subsection{Datasets}
We evaluate our methods on learning surrogate models of partial differential equations  (PDE) benchmark, and a more complex fluid dynamics prediction task.
\paragraph{Partial Differential Equations.} We include $4$ benchmark tasks in computational physics. The goal is to predict the spatial solution fields of $2$ PDEs, including Heat and Poisson's equations \cite{olsen2011numerical}. The ground-truth data is generated from the numerical solver. High-fidelity and low-fidelity examples are generated by solvers running with dense and coarse meshes, respectively. The output dimension is the same as the flattened mesh points. For both Heat and Poisson's equation with two-fidelity setting, they have $16 \times 16$ meshes at low fidelity level and $32 \times 32$ meshes at high fidelity level. For three-fidelity setting, they both have additional $64 \times 64$ meshes at the highest fidelity level. We calculate the relative cost of querying at each fidelity level $c_k$ based on the averaged computation time for data generation. We always set $c_1 = 1$ as a reference.

\paragraph{Fluid Simulation.} We also test \ours{} on a more challenging fluid dynamics simulation task. This computationally challenging simulation is based on the Navier-Stokes equation and the Boussinesq approximation \cite{holl2020learning}. We obtain the ground truth data by simulating the velocity field of smoke dynamics in a $50\times50$ grid. Initially, a static incompressible smoke cloud of radius $5$ is placed at the lower center of the domain together with a consistent inflow force is applied to the center at the initial position of the smoke. The inflow force varies in magnitude and direction for different scenarios. The two-dimensional input controls the magnitude of the inflow force at $x$ and $y$ directions. The output is the first component of the velocity field by applying the inflow for $30$ time stamps. We simulated the low fidelity ground truth with a $32\times32$ mesh and high fidelity with a $64\times64$ mesh. 

\subsection{Experiment Setup}

We consider two groups of experiments:
\begin{itemize}
    \item \textbf{Passive Learning}: model accuracy and robustness test by comparing the performance between \ours{} versus other baseline models using the entire training dataset. 
    \item  \textbf{Active Learning}: budget-constrained batch multi-fidelity active learning comparison between \ours{} with the MF-LIG acquisition function versus other multi-fidelity active learning frameworks.
\end{itemize}

For passive learning, we evaluate the performance of our model under three settings: nested, non-nested, and full. Let $\mathcal{X}_1$ and $\mathcal{X}_2$ to be two training input sets at $2$ fidelity levels. The ``full'' setting means that $\mathcal{X}_1 = \mathcal{X}_2$ and both sets have a large number of scenarios uniformly distributed in the input space, mimicking the final and convergent stage of active learning. The ``nested'' setting means that $\mathcal{X}_2 \subset \mathcal{X}_1$ and the ``nonnested'' settings means that $\mathcal{X}_1 \land \mathcal{X}_{q} = \mathcal{X}^{r}$, where $\mathcal{X}^{r}$ includes the inputs for the reference set. These two settings are used to mimic the early stage of active learning where the number of low-fidelity data points is much larger than the high-fidelity data points. We use these three settings to test the robustness of \ours{} and other baselines. For comparison, we consider state-of-the-art baselines for multi-fidelity surrogate modeling, including DMFAL~\cite{li2020deep}, NARGP~\cite{perdikaris2017nonlinear}, and MFHNP~\cite{wu2022multi}.


For active learning, we use the same $8$ uniformly sampled data points across all fidelity levels as the reference data for initial training. We run $25$ iterations and at each iteration, the active learning framework queries the simulator for the input with the highest acquisition function score until it reaches the budget limit of $20$ per iteration. We compare our method against DMFAL \cite{li2020deep}, BMFAL-Random \cite{li2022batch}, BMFAL \cite{li2022batch} and MF-BALD \cite{gal2017deep} as baselines, using the same hyperparameter settings as in the literature.

For both passive and active learning, we randomly generate $512$ data points as the test set for $4$ benchmark tasks and $256$ data points as the test set for fluid simulation.
We use the normalized Root Mean Squared Error (nRMSE) to measure prediction performance at the highest fidelity level, as our goal is to mimic the dynamics at the highest fidelity level. All experiment results are averaged over $3$ random runs. 

Our code is available at \href{https://github.com/Rose-STL-Lab/Multi-Fidelity-Deep-Active-Learning}{https://github.com/Rose-STL-Lab/Multi-Fidelity-Deep-Active-Learning}.


\begin{figure*}[t!]
  \centering
  \includegraphics[width=1.\linewidth,trim={0 0 0 0}]{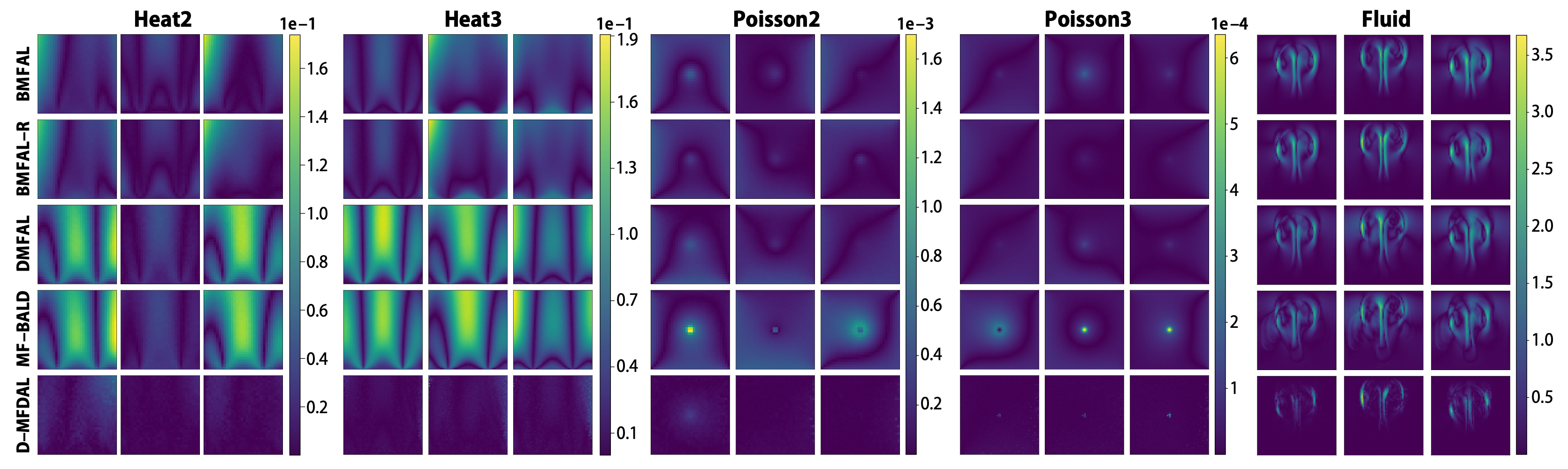}
  \caption{Prediction residual visualizations at the highest fidelity level for \ours{} and 4 baselines for Heat equation and Poisson's equation simulation with two and three fidelity levels, fluid simulation with two fidelity levels. For each simulation scheme presented, we randomly select three samples to visualize. Better performance is indicated by a darker color.
}
\label{fig:visualization}
\end{figure*}

\begin{figure*}[t!]
    \centering
    \includegraphics[width=1.\linewidth,trim={0 20 0 20}]{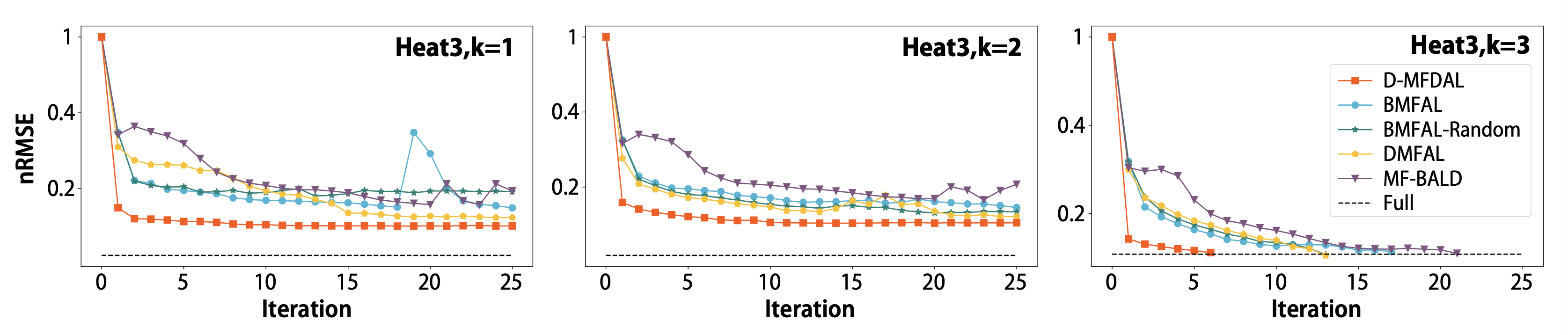}
    \caption{Active learning performance comparison for Heat3 simulation at three fidelity levels. Performance is measured at each fidelity level. $k$ represents the fidelity level. \ours{} outperforms the baselines across all fidelity levels.
    }
    \label{fig:fig4}
\end{figure*}

\subsection{Experimental Results}

\paragraph{Passive Learning Performance.} We test the passive learning performance of \ours{} and baselines across $5$ tasks and $3$ settings. The results are shown in Table \ref{tb:offline}. It can be seen that our model consistently outperforms all baselines across all settings and tasks. Furthermore, \ours{} performs particularly well under challenging nested and disjoint settings where the number of training data available at the highest fidelity level is limited. For example, in the complex fluid simulation, we find \ours{} with only $8$ data points at the high fidelity level under the nested setting outperforming all other baselines in the full setting.

The results show that \ours{} is capable of utilizing the information from the low fidelity levels to make good predictions at the highest fidelity level.  \ours{} is also quite robust as it almost has the best model performance under all three representative active learning settings. These advantages show that \ours{} is suitable for Bayesian active learning throughout the training process.


\paragraph{Active Learning Performance.} Figure \ref{fig:fig3} shows the nRMSE versus the number of iterations in active training. Our proposed \ours{} with MF-LIG always has the best nRMSE performance throughout the active learning process.
Furthermore, \ours{} converges to offline performance iterations faster than all other baselines for the Poisson2, Poisson3, Heat3 and Fluid experiments. 
Figure \ref{fig:visualization} is the visualization of prediction residuals for \ours{}, as well as $4$ other baselines. We visualize the
residual between the predictions and the truth to highlight the performance difference across $5$ datasets. A higher residual value indicates lower accuracy. We randomly select $3$ samples from the test set for each task.
It can also be found that \ours{} with MF-LIG outperforms other baselines as it successfully predicts the true patterns among all $15$ samples.






\paragraph{Ablation Study.} In Figure \ref{fig:fig4}, we compare active learning performance at $3$ fidelity levels on the Heat3 dataset. We find that the performance of \ours{} is always the best at each fidelity level, although the MF-LIG is designed to optimize the surrogate modeling performance at the highest fidelity level. Specifically, we find that the performance gap between \ours{} and the other baselines is consistently evident across all active learning iterations and fidelity levels. It shows one of the other advantages of our proposed \ours{}. That is, we can utilize the data at the high fidelity level to reversely improve the model performance at the low fidelity level. Although it is not the goal to improve surrogate modeling performance at lower fidelity levels in our tasks, it makes \ours{} flexible to be applied to general setups such as multi-task surrogate modeling where multiple tasks are considered. 

%% file: sections/conc.tex


To conclude, we design a multi-fidelity deep active learning framework, \ours{}, to learn functional relationships across multiple fidelity levels. \ours{} disentangles the individual latent representations, separating them into global and local terms to tackle issues of error propagation and overfitting. We design a unified ELBO over the joint distribution across all fidelity levels to serve as the training loss and include a multi-fidelity regularization term to infer the global representations across different levels of fidelity. Additionally, we  generalize the acquisition function, latent information gain, used in Bayesian active learning for NP-based models to multi-fidelity settings and  design an efficient algorithm for budget-constrained batch active learning. We conduct extensive empirical evaluations on several benchmark studies and complex spatiotemporal simulations to demonstrate the superior performance of our proposed \ours{} for both passive learning and active learning. For future work, we plan to extend this method for multi-task active learning. 